\documentclass[conference]{IEEEtran}

\usepackage{amsmath}
\usepackage{amssymb}
\usepackage{algorithm}
\usepackage{algpseudocode}
\usepackage{graphicx}

\newcommand\etal{\emph{et~al.}}
\newcommand\ie{\emph{i.e.}}

\newcommand\eg{\emph{e.g.}}

\begin{document}
\title{Detecting Curve Text with Local Segmentation Network and Curve Connection}

\author{
\IEEEauthorblockN{Zhao Zhou, Hao Ye, Luhui Chen, Yingbin Zheng}
\IEEEauthorblockA{Videt Tech Ltd., Shanghai, China}
}

\maketitle

\begin{abstract}
Curve text or arbitrary shape text is very common in real-world scenarios. In this paper, we propose a novel framework with the \emph{local segmentation network} (LSN) followed by the \emph{curve connection} to detect text in horizontal, oriented and curved forms. The LSN is composed of two elements, \ie, proposal generation to get the horizontal rectangle proposals with high overlap with text and text segmentation to find the arbitrary shape text region within proposals. The curve connection is then designed to connect the local mask to the detection results. We conduct experiments using the proposed framework on two real-world curve text detection datasets and demonstrate the effectiveness over previous approaches.
\end{abstract}

\IEEEpeerreviewmaketitle

\section{Introduction}
\label{sec:intro}

Text is probably the most important way for humans to communicate and express information. With the ubiquitous image capture devices, a huge amount of scene text images are produced, which brings a great demand for automatic text content analysis. The extracted text information also helps to understand the context of the whole images. As one of the most fundamental task for text analysis, scene text detection is to identify text regions of the given scene text images, which is also an important prerequisite for many multimedia tasks, such as image understanding and video analysis.

\begin{figure}[t]
\centering
\includegraphics[width=.99\linewidth]{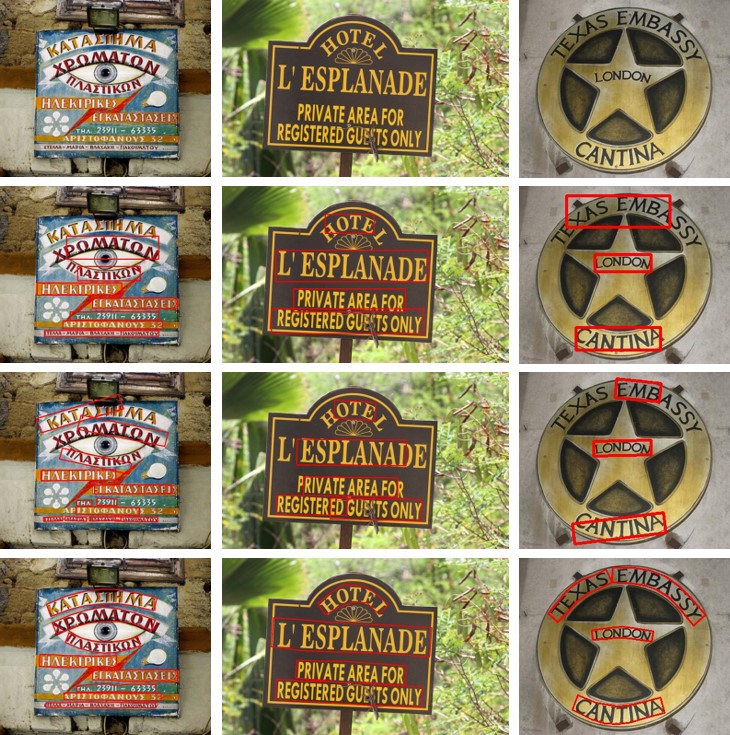}
\caption{Example text images in natural scenes (Row 1) and text detection results by different methods: CTPN~\cite{tian2016detecting} (Row 2), EAST~\cite{zhou2017east} (Row 3), and the proposed method (Row 4).}
\label{fig:sample}
\end{figure}

With the development of convolutional neural networks, there have been many attempts on text detection in natural scenes and great progress are achieved in recent years. The early attempts to detect text are with annotations of horizontal texts (\eg, in \cite{tian2016detecting}) and the approaches for arbitrary-oriented scene text detection are also proposed (\eg, in \cite{zhou2017east,liu2017deep,ma2018arbitrary}). However, in the real-world scenarios, there are still many text regions with irregular shapes, such as the curve words or logos, and a few of such sample images are shown in Fig. \ref{fig:sample}. It is still very challenging to detect these regions with different shapes.

\begin{figure}[t]
\centering
\includegraphics[width=\linewidth]{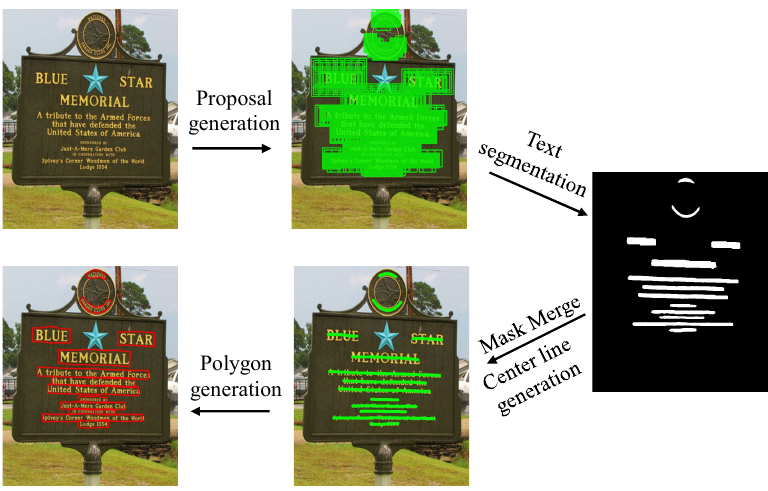}
\caption{Curved text detection pipeline of the proposed framework.}
\label{fig:pipeline}
\end{figure}

In order to detect text in horizontal, oriented and curved forms, we propose in this paper a novel framework incorporating the local segmentation network (LSN) and the curve connection and the pipeline of text detection is illustrated in Fig. \ref{fig:pipeline}. The LSN is designed with two functionality, \ie, proposal generation and text segmentation. We chose the ResNet-50~\cite{he2016deep} as the backbone of LSN. The horizontal rectangle proposals are generated based on multiple feature maps for the adaption to the different scales of the text. These detected proposals are small local regions but with high overlap with ground-truth text. Text segmentation is used to find the arbitrary shape text region within proposals by an ROI-Align~\cite{he2017mask} to produce the same size of the features from different anchors and a segmentation subnet to fine-tune the text area. The curve connection is then designed with center line generation and text polygon generation to connect the local mask to the detection results. We demonstrate a considerable improvement for the curve connection over those regressed directly from LSN. To evaluate our proposed framework, we report the evaluations on the recent proposed curve text detection datasets, \ie, CTW1500~\cite{yuliang2017detecting} and Total-Text~\cite{ch2017total}, and we compare with several recent approaches. 

The contributions can be summarized as follows.
\begin{itemize}
\item We propose a novel neural network architecture combing horizontal rectangle proposals and arbitrary shape text segmentation to perform curve text detection. Our framework can generate multi-scale proposals where only a small number of curve annotations are needed.
\item We also propose novel strategies for the curve connection of local text segments to improve the performance of long text words and arbitrary shape text detection.
\item We apply our framework to two real-world text detection datasets and the experiments show the effectiveness of the proposed framework compared to baseline metods.

\end{itemize}

The rest of this paper is organized as follows. Section \ref{sec:related} briefly reviews the literature of scene text detection. Our proposed framework is introduced in Section \ref{sec:framework} and the details are described in Section \ref{sec:lsn} and \ref{sec:cc}. In Section \ref{sec:exp}, we demonstrate the quantitative study on the benchmarks. Finally, We conclude our work in Section \ref{sec:conclusion}.

\section{Related Work}
\label{sec:related}

Scene text detection has drawn growing attention from computer vision communities in recent years. With the astonishing development of object detection, the state-of-the-art frameworks such as Faster-RCNN~\cite{ren2015faster} and SSD~\cite{liu2016ssd} have been widely applied to text detection field. However, compared with object detection that only needs general localization for objects, scene text detection requires precise positioning for characters. Therefore, many remarkable methods based on object detection for scene text detection have been proposed. These methods focus on more precise positioning for characters and can be roughly classified into two categories, \ie, the anchor-based methods and the link-based methods.

\vspace{0.08in}
\noindent\textbf{Anchor-based methods}. The horizontal box in object detection was widely used in document text detection but the orientation of box is various when facing scene text detection. Ma \etal~\cite{ma2018arbitrary} proposed a multi-oriented scene text detection approach by generating six-orientations anchors at each point of the feature map. Quadrilateral anchor was introduced in \cite{liu2017deep} to detect text with tighter quadrangle. Zhou \etal~\cite{zhou2017east} combined these representations together and proposed an efficient detector using a single fully convolution network with two branches. Liu \etal~\cite{yuliang2017detecting} applied 14 landmark points to represent curved text flexibly. These methods focused on pursuing a tighter boundary of text but they were limited due to the templates that lack variability to cover text with extremely aspect ratio, such as long text or non-quadrilateral text.

\vspace{0.08in}
\noindent\textbf{Link-based methods}. Link-based methods are more robust when facing scene with long text or non-quadrilateral text. Tian \etal~\cite{tian2016detecting} introduced CTPN using LSTM~\cite{graves2005framewise} to link several text proposals. Shi \etal~\cite{shi2017detecting} proposed the SegLink framework that decomposes the text into segments and links and links text segments together. And Tian \etal~\cite{Tian2015Text} introduced a graph method called Min-Cost Flow to link single characters. However, these  methods are based on a strong prior by restricting all the proposals to link should lie in a line. This hypothesis, in a manner, makes the problem easier and tractable but can not handle curved text.

Recently, several approaches have been proposed to detect text with form of arbitrary shapes and the curve text datasets were provided for research. \cite{yuliang2017detecting} proposed a polygon-based curve text detector (CTD) which can directly detect curve text without empirical combination. The framework of TextSnake~\cite{long2018textsnake} considered a text instance as a sequence of ordered disks. To deal with the problem of separation of the close text instances, Li \etal~\cite{li2018shape} designed the PSENet by a progressive scale algorithm to gradually expands the pre-defined kernels. Different from previous methods, our local segmentation network combines the advantages of link-based methods and anchor-based methods using proposals to rough locate text and get accurate boundary by text segmentation. We also demonstrate the effectiveness of the proposed framework through experiments on the recent curve text detection datasets.

\section{Framework}
\label{sec:framework}
The pipeline of the proposed framework is illustrated in Fig. \ref{fig:pipeline}. Each text can be represented by a bunch of text segments. For each segment, it has a coarse boundary represented by a square and a fine boundary represented by a mask. The former can fast classify into foreground/background proposals by a detection task, while the latter is used to get a tighter boundary of text. When faced with long text or curved text, segments will be linked together through their masks. After using the curve connection operations including the mask merging, center line generation, and polygon generation, a more smooth and robust boundary of text can be obtained.

\section{Local Segmentation Network}
\label{sec:lsn}

\begin{figure}[t]
  \centering
  \includegraphics[width=\linewidth]{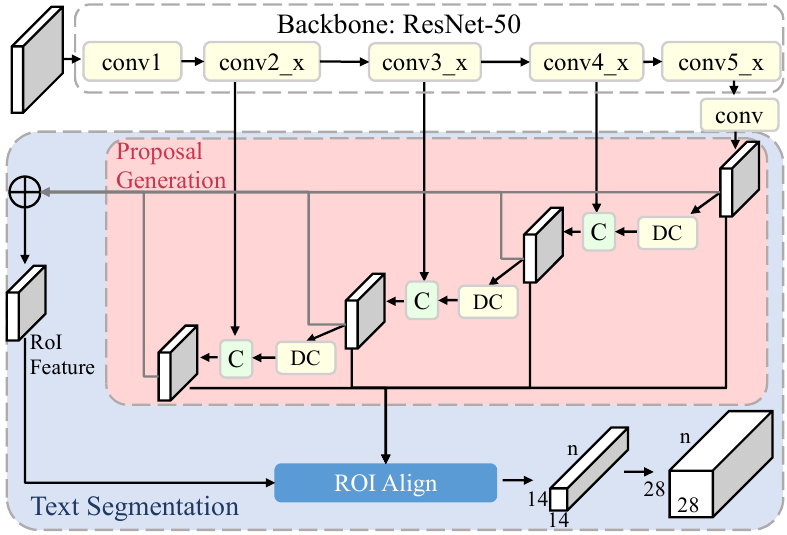}
  \caption{The architecture of Local Segmentation Network. C and DC indicates the concatenation operation and the de-convolution layer.}
  \label{fig:network}
\end{figure}

\subsection{Network Structure}
Our network is shown in Fig. \ref{fig:network}. Here we choose ResNet50 \cite{he2016deep} as the backbone to capture more detailed information. We remove the last fully-connected layer in ResNet50 and concatenate extra convolution layer to get deeper features with larger receptive fields. We feed the feature maps after each stage to the feature merging network. For each feature merging network, we apply a convolutional layer with $1\times1$ kernels to make two merge features have the same dimensions. In addition, we use 4 different stride features for the class square. These features can provide rich information and different receptive fields for robust detection. In order to get both fine and coarse feature, these 4 features are concatenated as the ROI feature. After the ROI feature, we use ROIAlign~\cite{he2017mask} to get same size feature with different size of the square. Finally, we upsample the features to predict text mask.

\begin{figure}[t]
  \centering
  \includegraphics[width=.8\linewidth]{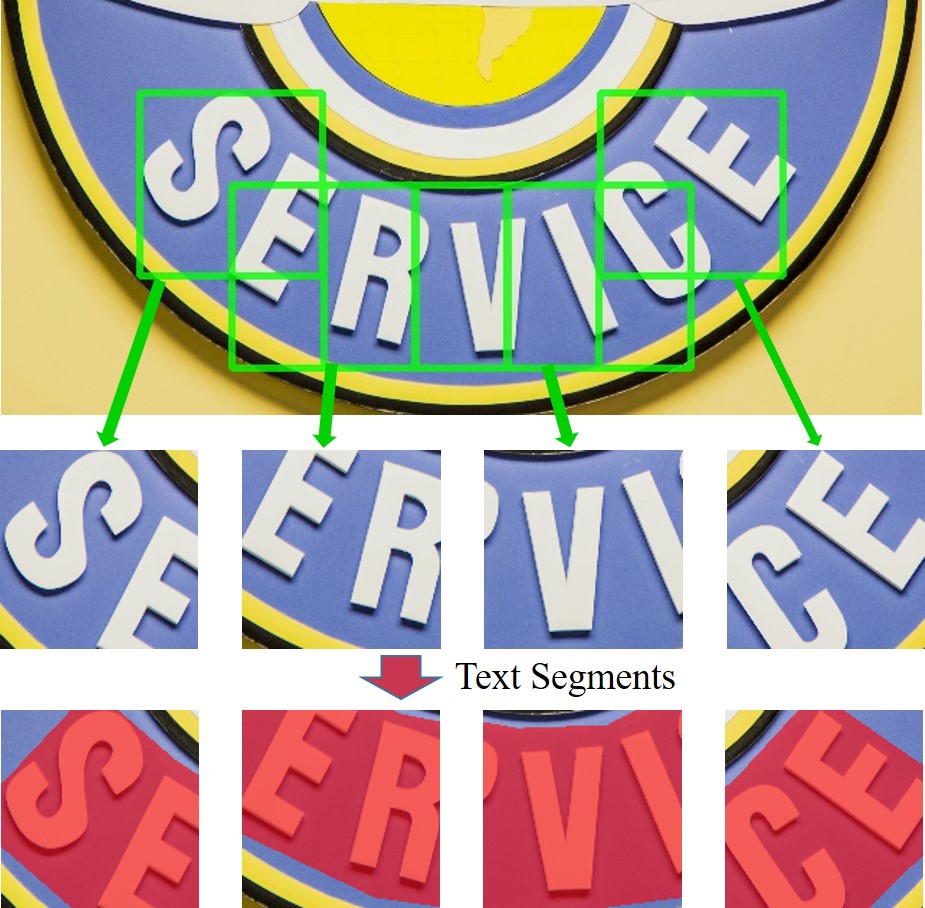}
  \caption{Sample image region with text segmentation after proposal generation.}
  \label{fig:classAndSeg}
\end{figure}

\subsection{Proposal Generation}
As is shown in the Fig. \ref{fig:classAndSeg} top, a text region can be covered by several overlap square. We define each square through $(x,y,l)$. Here $x, y$ represents the center coordinate of the square. Actually, we consider the size of feature map extracted from network is $w$ and $h$. The relation between the coordinate of square and the feature map can be formulated as the following equations:
\begin{equation}
\begin{aligned}
x_{i} &= &i \times \mathrm{stride}, i \in \{0,1...w-1\} \\
y_{j} &= &j \times \mathrm{stride}, j \in \{0,1...h-1\}
\end{aligned}
\end{equation}
$l\in \{ s \times k | s = 8,16,32,64.~k = 2,2.5,3,3.5\}$ is the width of the square that can cover nearly all the scale of text. The network only produce 2 channels for text classification. Each predict feature point has 4 different anchor sizes.

\subsection{Text Segmentation}
In order to get a tight boundary of the text region, we predict a text mask for each positive square after the text classification stage. ROIAlign is used to get the same-size feature from various size of ROI features. Then the extracted features are concatenated and upsampled to get the final mask result. The result of the processing of text segmentation above is shown in Fig. \ref{fig:classAndSeg} bottom.

\begin{figure}[t]
  \centering
  \includegraphics[width=.99\linewidth]{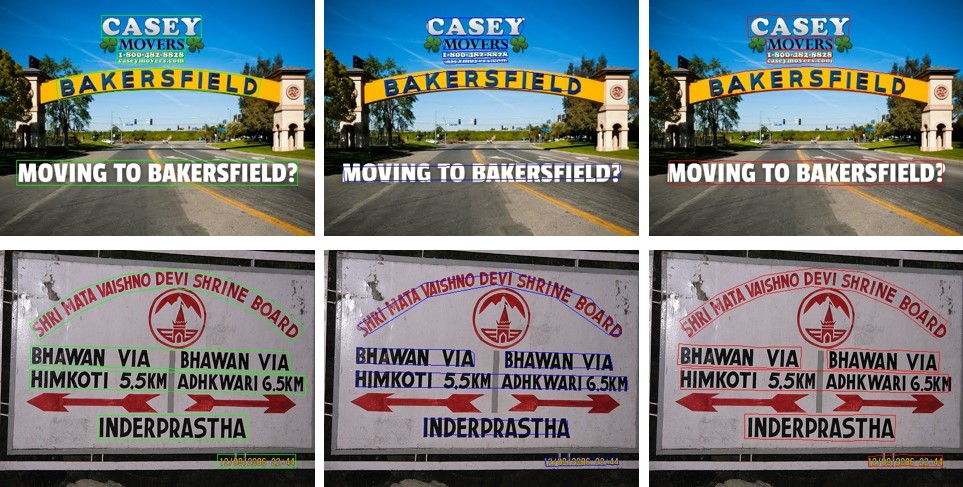}
  \caption{Effect of segment merging strategy. Left: original images with ground truth; Middle: results by neighboring segment connection; Right: results by curve connection.}
  \label{fig:cc}
\end{figure}

\subsection{Loss Function}
To train the LSN, the loss function is formulated as:
\begin{equation}
\begin{aligned}
L = \lambda_1 L_{\text{cls}} + \lambda_2 L_{\text{segment}},
\end{aligned}
\end{equation}
where $L_{\text{cls}}$ and $L_{\text{segment}}$ represent the classification loss and segmentation loss for text instances respectively. $\lambda_1$ and $\lambda_2$ balances the importance between  $L_{\text{cls}}$ and $L_{\text{segment}}$.

Focal loss~\cite{lin2018focal} is employed for classification term $L_{\text{cls}}$,
\begin{equation}
\begin{aligned}
 L_{\text{cls}} = \sum_{i \in {1,2,3,4}}{\text{FocalLoss}(p_{cls}(i),g_{cls}(i))}\\
\text{FocalLoss}(p_{t})= - \alpha_{t}(1-p_{t})^\gamma log(p_{t})
\end{aligned}
\end{equation}
where  $p_{cls}(i)$ and $g_{cls}(i)$ is the $i_{th}$ feature predicts and ground truth, and $\alpha_{t}$ and $\gamma$ is hyper-parameters  in focal loss.

For the segmentation term $L_{\text{segment}}$, we select $smooth_{L_1}$ loss to get the score of each pixel in square region and the definition is as follows:
\begin{equation}
\begin{aligned}
\label{Lseg}
 L_{\text{segment}} = smooth_{L_1}(p_{segments},g_{segments})
\end{aligned}
\end{equation}
\begin{equation}
\text{smooth}_{L_1}(x)=\left\{
\begin{array}{ll}
0.5x^2 & \text{if~} |x| < 1\\
|x| - 0.5 & \text{otherwise}
\end{array} \right.
\end{equation}

In our experiments, the hyper-parameters $\lambda_1$ and $\lambda_2$ are set to 1, the weights $\alpha_{t}$ and $\gamma$ in focal loss are set 0.25 and 2 respectively.

\renewcommand{\algorithmicrequire}{ \textbf{Input:}}
\renewcommand{\algorithmicensure}{ \textbf{Output:}}
\begin{algorithm}[t]
\caption{Mask merging for a given image.}
\begin{algorithmic}[1]
\Require
Predict mask set $S$; threshold $s_1$, $s_2$
\Ensure
Merged mask queue $Q$

\State$ Q = \{\} $
\For {each predict mask $p \in S$}
\If {$p_{i, j} > s_1$}
\State $p_{i, j} = 0$
\Else
\State $p_{i, j} = 1$
\EndIf
\State overlap list $L  = \{\}$
\For {each item mask $m_i \in Q$}
\State merged mask $m = m_i \cup p $
\State overlap ratio $r = area_{m} / \min (area_{m},area_{p})$
\If {$  r >  s_2 $}
\State insert $i$ into $L$
\EndIf
\EndFor
\State $m=p$
\For {each $i \in L$}
\State $ m = m \cup Q_{L[i]} $
\State delete $Q_{L[i]}$
\EndFor
\State insert $m$ to $Q$
\EndFor
\State \Return $Q$
\end{algorithmic}
\end{algorithm}

\begin{algorithm}[t]
\caption{Text polygon generation.}
\begin{algorithmic}[1]
\Require
Merged mask set $M$ for a given image
\Ensure
Text polygon set $P$ for the image
\State $P = \{\}$
\For {each proposal mask $ m\in M$}
\State Polygon $p$ = []
\State choose $n$ positive pixels
\State use principal curve to find curve center line
\State choose 7 points $p_i$ ($i=0...6$) to represent center line
\For {$i \in [0,6]$}
\State generate circumscribed rectangle of the area between $p_i$ and $p_{i+1}$
\State regard rectangle points as polygon points
\State insert the left two points into $p$
\EndFor
\State insert the $p$ into $P$
\EndFor
\State \Return $P$
\end{algorithmic}
\end{algorithm}

\section{Curve Connection}
\label{sec:cc}

As the text segments are generated, a merging strategy for the segments is needed to have the overall detection results. A straightforward approach is to connect the neighboring segments. However, the boundary cannot be properly detected, resulting in a large precision loss regarding the performance (see Fig. \ref{fig:cc}). In this section, we design the curve connection with the components of mask merging and text polygon generation.

\subsection{Mask Merging}

For each predict mask, we set a threshold $s_1$ to define where is the fine region of text. When the pixel predict score is higher than $s_1$, we set the pixel value 1 otherwise 0. In order to merge all boxes in the same text region, we design the mask merging approach and the pseudo-code is shown in Algorithm 1. First, we use a queue $Q$ store the regions who are separate from each other. And a new region will compare with any of the regions on $Q$. If the overlapping mask of these two regions take the proportion of the smaller region mask is above threshold $s_2$, these regions should be merged. After comparing with all of the regions on $Q$, we merge all of the regions which should be merged including the new region. The last two steps is repeated until all of the regions are operated.

\subsection{Text Polygon Generation}
A few text regions are obtained after the mask merging process. We choose $n$ positive pixels in each text region and use the principal curve~\cite{hastie1989principal} to regress the curve center line. Seven points are chosen from the center line. For each pair of the points that are adjacent in center line, we use center point of two points as a rectangle center and generate a circumscribed rectangle of the area where are the text region between the pare center points. We will get a polygon of the text region by repeating these steps. Detail of the algorithm is shown in Algorithm 2.

\section{Experiments}
\label{sec:exp}
\subsection{Datasets}
We evaluate our approach with two recent proposed curve text detection benchmarks, \ie, the CTW1500~\cite{yuliang2017detecting} and the Total-Text dataset~\cite{ch2017total} dataset.

\vspace{0.08in}
\noindent\textbf{CTW1500}~\cite{yuliang2017detecting} contains 1500 images (1000 train and 500 test). Each text instance annotation is a polygon with 14 vertexes to define the text region at the level of the text line. The text instances include both inclined texts as well as horizontal texts.

\vspace{0.08in}
\noindent\textbf{Total-Text}~\cite{ch2017total} contains not only horizontal and multi-oriented text instances but also the curved text. The dataset is consists of 1255 training images and 300 testing images. The images are annotated at the level of the word by a polygon with $2N$ vertices ($N \in  \{2,..., 15\}$).

\subsection{Implementation Details}

\noindent\textbf{Proposal label generation.} Assume the feature size is $W \times H$, our approach will generate $W \times H \times 4$ default anchors. For the label of each text polygon, two rules are defined to judge whether the default anchor is positive: 1. the center point of the default anchor is in ground truth and the height of default anchor is not great than 1.8 times of ground truth polygon height; 2. one of top two points in default box outside of ground truth polygon and the button is same as top. If both conditions are met, we consider it to be a positive anchor, otherwise, it is negative.

\vspace{0.08in}
\noindent\textbf{Text segment mask generation.} In order to separate the text instances that are very close to each other when generating text masks, we chose 50\% region of ground truth polygon as a strong true region and set their scores to 1. The rest 50\% regions are considered as the weak true region with a score of 0.1. The other regions are with the background score of 0.

\vspace{0.08in}
\noindent\textbf{Data augmentation.}
The images are randomly rotated with 30 degrees, aspect ratios are set from 0.33 to 3 and randomly reverse image left and right. After these steps, random crop invert and blur are randomly adjusted. In order to ensure the cropped image have complete ground truth polygon, we only randomly crop the most left box to image width and apply it to the other three directions.

\vspace{0.08in}
\noindent\textbf{Model training.}
Our method is implemented in Pytorch.
We use Adam optimizer as our learning rate scheme. During the training stage, we chose 200 positive anchors for each feature as an input of ROI-Align, and at the testing stage we set $s_3 = 0.4$ as the positive square threshold and the maximum number of the positive square to ROI-Align is 2000.

\begin{figure}[t]
  \centering
  \includegraphics[width=.99\linewidth]{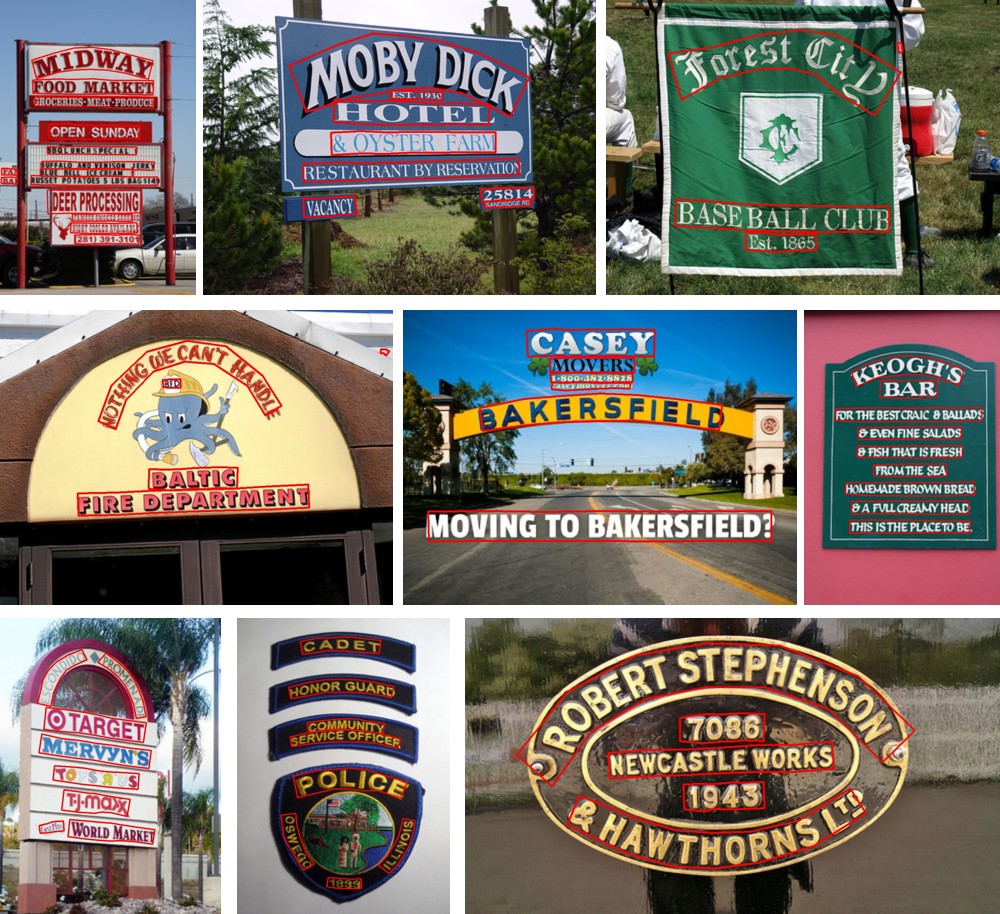}
  \caption{Text detection results on CTW1500.}
  \label{fig:ctw_results}
\end{figure}

\begin{table}[t]
\centering
\caption{Quantitative results of different methods on CTW1500. LSN+CC indicates the full framework, while LSN is the results without the curve connection.}
\label{table:ctw}
\begin{tabular}{c| c c c}
 \hline
 \textbf{Method} & \textbf{Precision} & \textbf{Recall} & \textbf{F-measure} \\
 \hline
 CTD~\cite{yuliang2017detecting} & 74.3 & 65.2 & 69.5 \\
 CTD+TLOC~\cite{yuliang2017detecting} & 77.4 & 69.8 & 73.4 \\
 SLPR~\cite{zhu2018sliding} & 80.1 & 70.1 & 74.8 \\
 TextSnake~\cite{long2018textsnake} & 67.9 & 85.3 & 75.6 \\
  \hline
 LSN  & 69.0 & 75.7 & 72.2 \\
 LSN+CC  & 83.2 & 78.8 & 80.8 \\
 \hline
\end{tabular}
\end{table}

\begin{figure}[t]
  \centering
  \includegraphics[width=.99\linewidth]{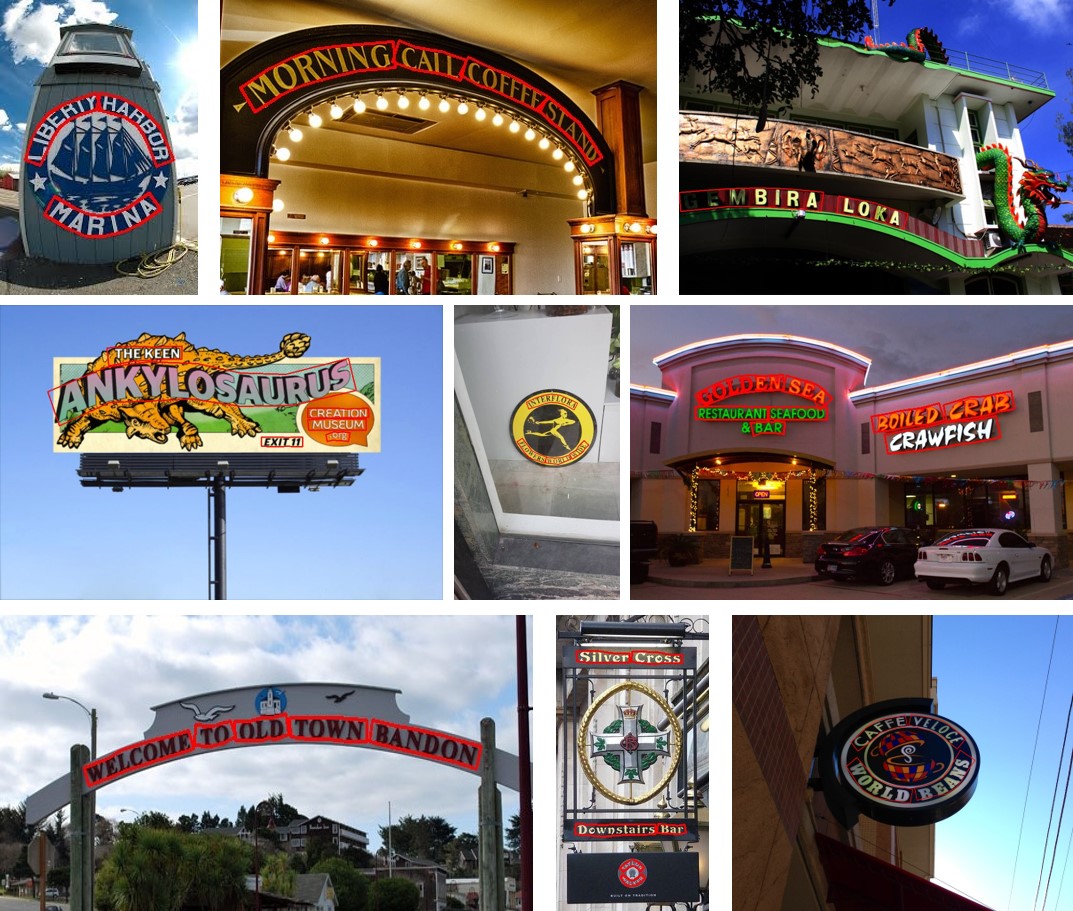}
  \caption{Text detection results on Total-Text.}
  \label{fig:tt_results}
\end{figure}

\begin{table}[t]
\centering
\caption{ Quantitative results of different methods on Total-Text.}
\label{table:ttext}
\begin{tabular}{c| c c c}
 \hline
 \textbf{Method} & \textbf{Precision} & \textbf{Recall} & \textbf{F-measure} \\
 \hline
 Total-Text~\cite{ch2017total} &40.0 &33.0 &36.0 \\
 Mask TextSpotter~\cite{yao2018mask} & 69.0 & 55.0 & 61.3 \\
 TextSnake~\cite{long2018textsnake} & 82.7 & 74.5 & 78.4 \\
 \hline
 LSN+CC & 82.4 & 76.9 & 79.5 \\
 \hline
\end{tabular}
\end{table}

\subsection{Results and Comparison}

\noindent\textbf{CTW1500.}
To evaluate our framework, we compare with different curve text detection methods: CTD and CTD+TLOC~\cite{yuliang2017detecting} by 14 landmark points to represent curved text, SLPR~\cite{zhu2018sliding} with sliding line point regression, and TextSnake~\cite{long2018textsnake} which represents curve regions as a sequence of ordered, overlapping disks. Table \ref{table:ctw} and Fig. \ref{fig:ctw_results} shows the results for text detection. It clearly shows that our proposed framework outperforms recent methods on CTW1500. The improvement of LSN+CC over LSN demonstrate that adding the curve connection can not only achieve a significant improvement of precision but also helps the recall.

\vspace{0.08in}
\noindent\textbf{Total-Text.}
Table \ref{table:ttext} demonstrate the comparison of our full framework with previous curve text detection approaches. We compare with Total-Text~\cite{ch2017total}, Mask TextSpotter~\cite{yao2018mask}, and TextSnake~\cite{long2018textsnake}. The substantial performance gains over the published works confirm the effectiveness of using the local segmentation network and then curve connection for the text detection task. Some detection results obtained on the benchmark are illustrated in Fig. \ref{fig:tt_results}.

\section{Conclusions}
\label{sec:conclusion}

We proposed a text detection framework for the arbitrary shape scene text. Initial curve proposals were generated with the text segmentation on the horizontal rectangle local-level proposals by the local segmentation network. By combining the text segments sent from the previous step, the proposals can be refined in terms of the text polygon with the curve connection algorithms. Experimental comparisons on CTW1500 and Total-Text showed the effectiveness of the proposed framework for curve text detection.

\bibliographystyle{IEEEtran}
\bibliography{total}

\end{document}